\documentclass[10pt,twocolumn,letterpaper]{article}

\usepackage{cvpr}
\usepackage{times}
\usepackage{epsfig}
\usepackage{graphicx}
\usepackage{amsmath}
\usepackage{amssymb}

\usepackage[breaklinks=true,bookmarks=false]{hyperref}

\cvprfinalcopy 

\def\cvprPaperID{****} 

\begin{document}

\title{Don't only Feel Read: Using Scene text to understand advertisements}
\author{Arka Ujjal Dey\\
IIT Jodhpur\\
{\tt\small dey.1@iitj.ac.in}
\and
Suman K. Ghosh and Ernest Valveny\\
Computer Vision Center, Dept. Ci\`{e}ncies de la Computaci\'o\\Universitat Aut\`{o}noma de Barcelona\\ 08193 Bellaterra (Barcelona), Spain \\
Email: sghosh,ernest@cvc.uab.es
}


\maketitle

\begin{abstract}
We propose a framework for automated classification of Advertisement Images, using not just Visual features but also Textual cues extracted from embedded text. Our approach takes inspiration from the assumption that Ad images contain meaningful textual content, that can provide discriminative semantic interpretetion, and can thus aid in classifcation tasks. To this end, we develop a framework using off-the-shelf components, and demonstrate the effectiveness of Textual cues in semantic Classfication tasks.  
\end{abstract}

\section{Introduction}


In the recent past deep CNNs have generated state of the art results in various computer vision tasks. While it started from character recognition ~\cite{lecun1998gradient}, the architecture has been successfully adapted to a whole range of allied task involving natural images, word images ~\cite{19wilkinson2016semantic}, as well as scene text images ~\cite{18jaderberg2016reading}. Alongside this with the advent of big data, deep learning has been applied in various Natutal Language Processing tasks, long term sequence learners like LSTM~\cite{2sundermeyer2012lstm}, and context encoders like CBOW ~\cite{3mikolov2013efficient} are being explored for language modeling tasks.
Sequential nature of text data allows it to be modeled by such encoders to provide semantic understanding of text data \cite{le2014distributed}. 

\begin{figure}
\includegraphics[width=80mm, height=70mm ]{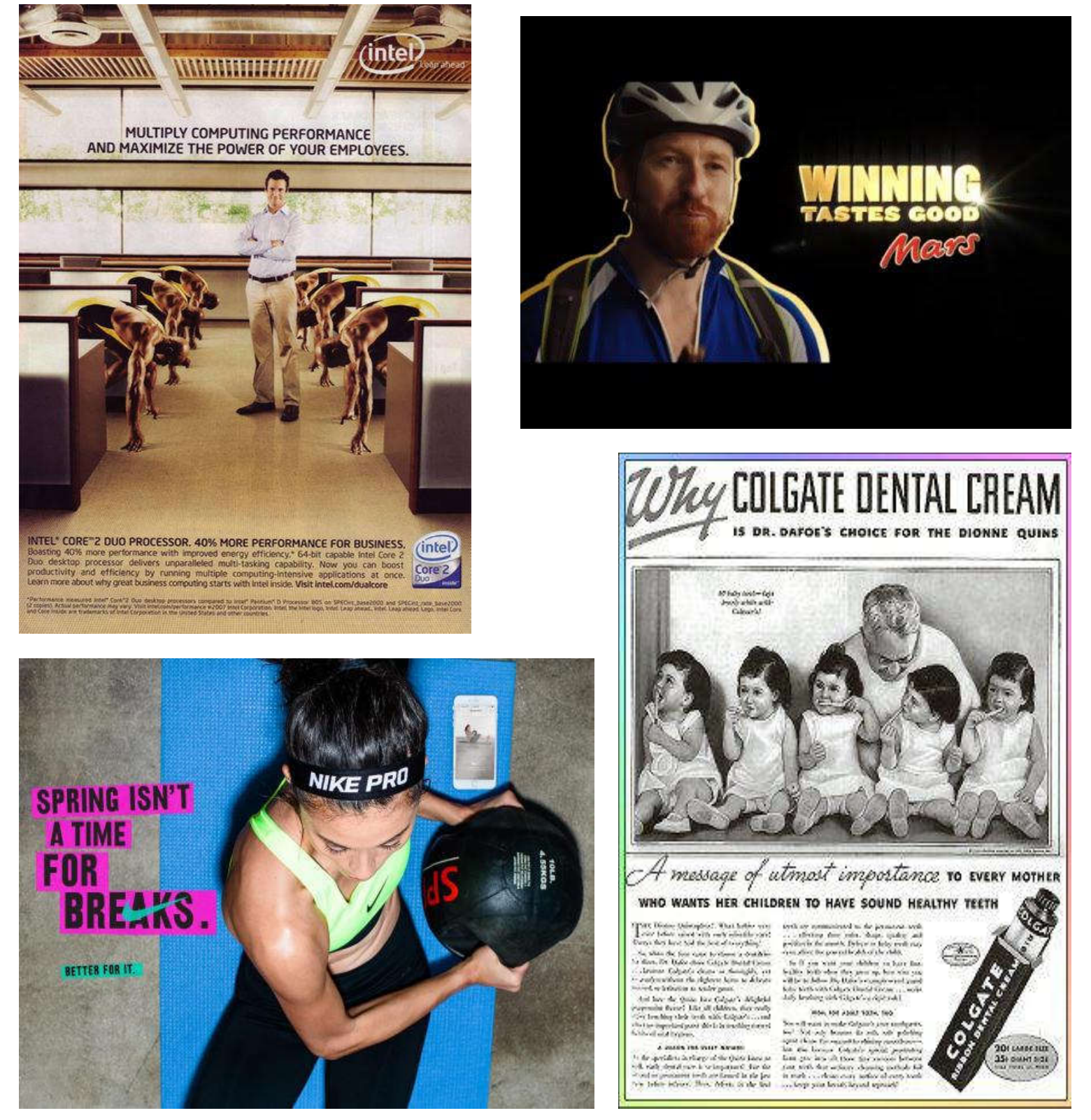}
\caption{Example Ad images, illustrating the complementary nature of text and visual cues. In some cases the visuals can be symbolic, but embedded text gives away the context[top-left, top-right], in other cases the visuals can be simple to understand but the text can be obtuse[bottom-left]. Further, the amount of text content can vary widely [top-right, bottom-right]  }\label{imgeg}
\end{figure}

This understanding text and images can be used  to solve more general AI problems like Image Captioning ~\cite{4kiros2014unifying}, Image Annotation, Visual Question Answering and Feature grounding ~\cite{5kottur2016visualword2vec}. Thus the interplay between text and image data  is necessary for all aforementioned tasks, 

In this context it is worth noting that images around us, apart from the visual semantic content, also contain a lot of embedded text (usually called scene text), which are provided for better human understanding of the images, whenever the image itself can not make the idea explicit. For example consider the image in Figure \ref{imgeg}. The content of the images is not clear without explicitly reading the text. 

In this work we build on this hypothesis that scene text (whenever available) plays very important role in understanding the image and thus we will use scene text as an extra cue and analyze its impact in in semantic image understanding. 
\section{Related Work}
In this section we briefly analyze the works which use two modalities to solve computer vision or AI problems.
Dealing with multimodal data poses two challenges, firstly on how to represent each of them, to effectively model the interplay and second modeling the interaction between them. 
The first step towards that is the generation of Global feature representation explictly or implicitly from local features.  
In case of images, Deep CNNs generate robust global features as the last fully connected layer activations, implictly from local CNN feature maps. For textual content on the other hand its global representation mainly depends on the presence or absence of structure or sequence. For structured text, like question/answers about an image, or sentence captions, LSTM-RNN language models are the obvious choice due to their superior ability in modeling the sequential data.
However for unstructured text like tags, or other Meta data, which is not sequential, the global feature is usually an aggregation (e.g. mean or averaging) of local semantic features. In some cases people have even used bag-of-words ~\cite{8clinchant2013textual} and fisher vectors ~\cite{9klein2015associating} for data aggregation to model the first order and second order statistics respectively. Once these Global textual and Visual Feature representations are computed, one has to model the interplay between them and this, depending on the end goal, one can have various ways.

For multimodal retrieval tasks usually the aim is to project both modalities into a common subspace – where they are comparable, CCA (Canonical Correlation Analysis) ~\cite{hardoon2004canonical} and its deep learning variant (DCCA) ~\cite{andrew2013deep} are the most useful techniques often used by these algorithms. 
However this does not generate any unified third representation, encoding both the text and image features and lacks interoperability between them. 



To effectively model the inter-relationship between them, generating one unified representation which can represent the data in higher semantic hierarchy, a feature fusion scheme is needed.
Early methods use simple techniques like element wise sum/product, concatenation of features etc. However this is not expressive enough to capture the rich and non-linear relation between features from two different views/modalities. Outer product based bilinear interaction schemes like MCB ~\cite{21fukui2016multimodal}, Mutan ~\cite{20ben2017mutan} allow for multiplicative interaction between the features, and are thus more suited. These schemes are used for image caption generation by leveraging the co-relation between text and image space at various level of granularity. In this work we will analyze the efficacy of both simple fusion schemes like average and concatenation and more generic outer join based schemes.

\section{Methodology}
\begin{figure}

\includegraphics[width=85mm, height=50mm ]{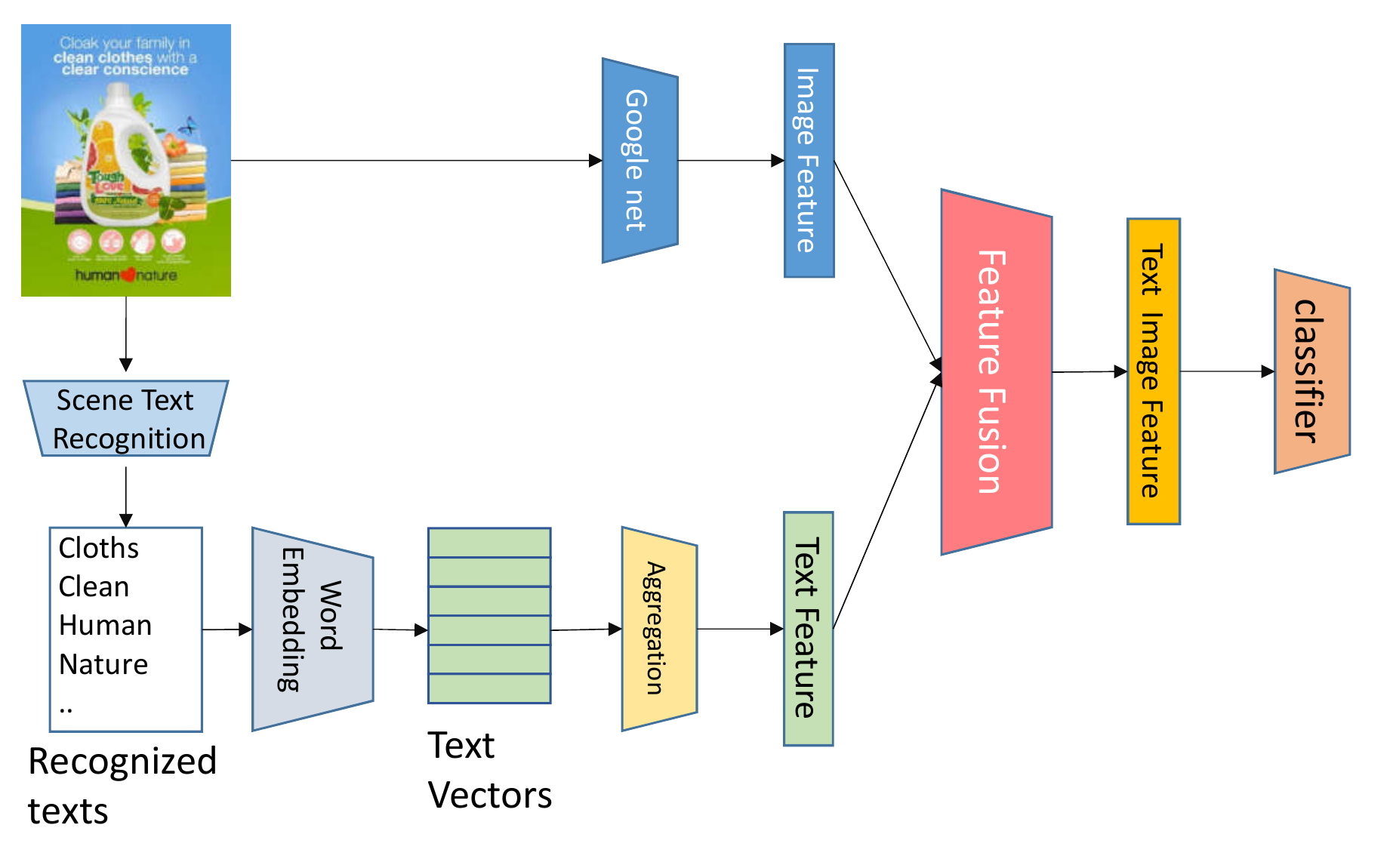}
\caption{Topic Classification framework}\label{framework}
\end{figure}

Though textual content in images is ubiquitous in our everyday life, in the form of newspaper, magazines, print ads, store fronts, street scenes, their use in solving general problem other that text understanding itself has not been studied much in the literature. In this work our aim is to use these textual data in order to understand the world around us.
In particular we deal with the problem of image classification into semantic topics, our  framework is shown in Figure \ref{framework}. We also apply our framework to visual question answering task from advertisement images. Our basic network is composed of three parts namely the scene text understanding part, image feature extraction and data fusion.

\subsection{Scene Text Understanding}\label{textdes}
Though embedded text (from images) can be a rich source of information about semantic understanding this text is not there in usual text (ASCII) format. Rather it is embedded within image pixels, thus to leverage this knowledge the first task is to extract this texts from the raw image.

Thus, the first task in such a pipeline is to extract the text from the images. However this is not simple task and in reality is an active area of research \cite{18jaderberg2016reading}.

In this work, as our goal is not to effectively extract text but to analyze the efficacy of text in understanding the images, we use an standard off-the-self model for text detection ~\cite{23liao2017textboxes} and recognition~\cite{18jaderberg2016reading} pipeline to detect and transcribe text. 


Once we obtain a list of text extracted from an image with corresponding confidence measures we embed these texts into a semantic vector space such that words with similar semantics are have similar vector representation. 
In our current experiments we have used the word2vec ~\cite{3mikolov2013efficient} semantic embedding, as this has been successfully used in different semantic understanding pipeline. 

Since the number of detected text varies widely from image to image in case of advertisement images, we limit ourselves to use only the top $k$ most discriminatory words according to tf-idf score. Experiment with different values of $k$ is presented in experimental results section ~\ref{textorimage}.  

We aggregate the $k$ corresponding word2vec vectors to generate a global text feature for a given image. The vector structure of the word2vec space, allows for simple vector sum to be a meaningful aggregation scheme. In experimental section we will show that using only text feature in this manner is comparable to image feature in understanding the semantics of the advertisement.


\subsection{Image Feature Extraction}
As the focus of our work is not to study the semantics of image features, which is a well studied topic we restrict our investigation only to standard CNN based feature extractors. In particular we use ~\cite{szegedy2015going} to extract global feature vector from every image.

\subsection{Combining Text and Image Feature, Data fusion}

\begin{figure}

\includegraphics[width=80mm, height=40mm ]{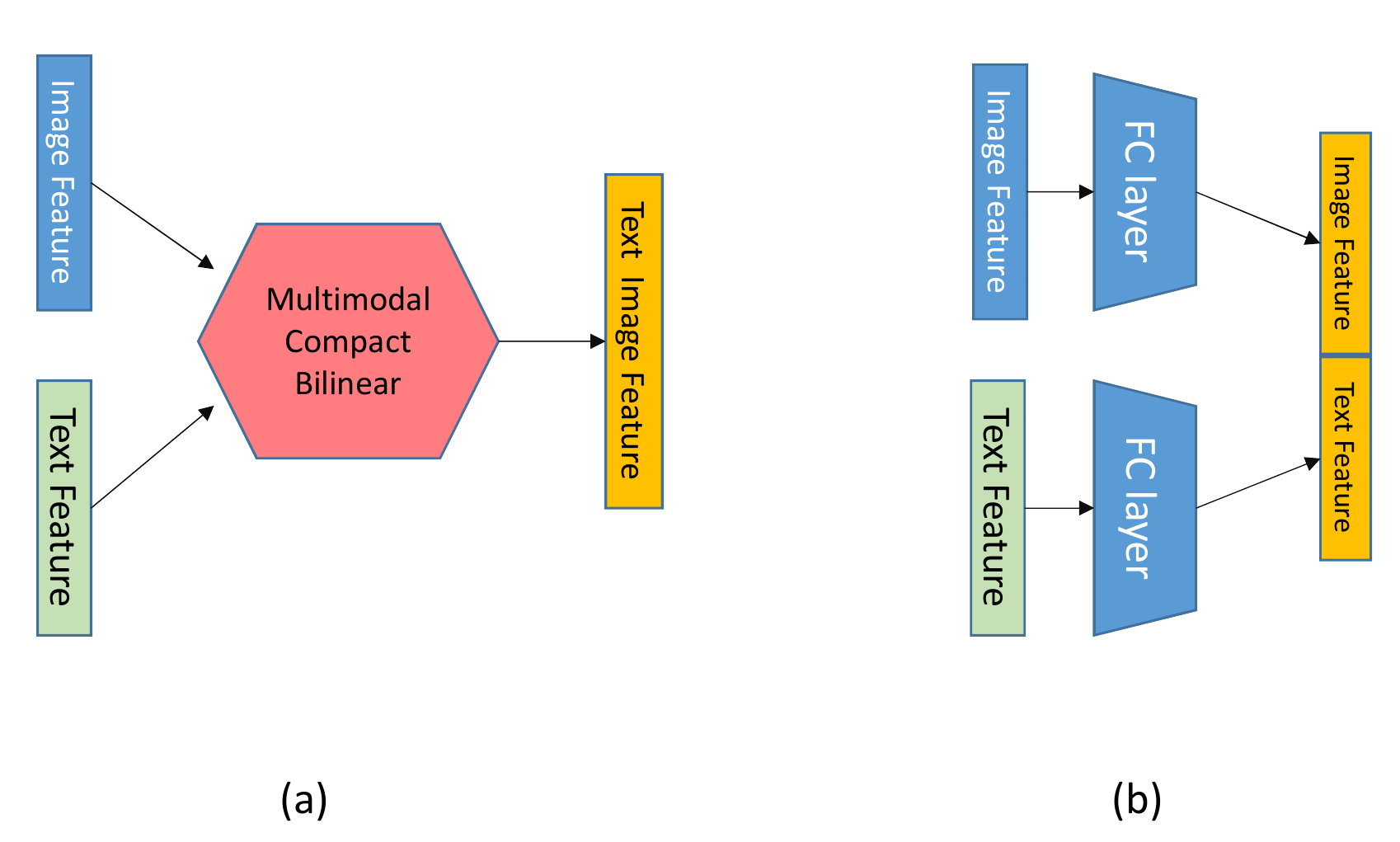}
\caption{Fusion schemes , (a) MCB , (b) Concat}\label{fusion scheme}

\end{figure}

In our current set of experiments, we have explored one simple and one generic data fusion scheme. For baseline we use a simple concatenation and then we learned outer product approximation as fusion schemes\ref{fusion scheme}.
We used a similar formulation like MCB ~\cite{21fukui2016multimodal}.
In particular we use $1024$ dimensional feature from last fully connected layer of googlenet \cite{szegedy2015going} and text feature as desribed in Sec. ~\ref{textdes}. Now Outer product between these two views is approximated. A low rank approximation is achieved by using a count sketch transforation.

In our experiments, we found outer product based on MCB \cite{21fukui2016multimodal} leads to better accuracy. 


\section{Experimental Results}
We evaluate the benefit of using text features, when attempting to understand Ad images. We argue, that careful selection of relevant texts, imply rich semantic information, which can aid the visual cues, and can result in better classification results. We use the Jaderberg ~\cite{18jaderberg2016reading} scene text recognition engine on top of bounding boxes generated by Textboxes ~\cite{23liao2017textboxes} to generate this transcription. 
However the nature of the Ad dataset, with varying amount of legible text, and Vocabulary limitations of our recognition engine, implied that we were not able to generate meaningful text features for every Ad image. To this end, we generated a cleaner dataset, consisting of 47000 images for which we are able to generate accurate transcriptions with $70\%$ confidence.
Our experiments are conducted on this dataset, where every image has some legible text leading to a meaningful text feature.

\subsection{Topic Classification Task}
In the Topic classification task, the objective is to classify an Ad image into 1 of 40 Topic classes. As shown in \ref{imgeg}, sometimes the visual content can be symbolic or metaphorical, while the text content is more straight forward to interpret or in other cases it could be that the text is misleading (eg. 'spring' , 'time' in the nike Ad), but the visual content is fairly straight. In conclusion, for Ad images both visual and textual cues and generate power features \ref{textorimage}, but can lead to complementary interpretations. This lead us to try fusion schemes, whereby we learn a feature set using both, the text and visual cues. As is demonstrated in 
\ref{textnimage}, fused features lead to better classification accuracy. Further, the multiplicative nature of MCB, allows for more interaction between the features than simple concatenation, and thus leads to a better performance. 

\begin{table}[h]
\small
\caption{Topic Classification accuracy using only Image Features and only embedded text features from k most significant text words .}\label{textorimage}
 \begin{tabular}{|c|c|c|c|c| }
 \hline
 Image & Text k=5 & Text k=10 &Text k=35 & Text k=100 \\ \hline
 45 &41 &41 &40 &40 \\ \hline
 
 \end{tabular}
\end{table}

\begin{table}[h]
\small
\caption{Topic Classification accuracy using fusion of Image Features and embedded text features.}\label{textnimage}
 \begin{tabular}{| l |l| l | c |}
 \hline
 Fusion  & Image Text k=5 & Image Text k=35  & Image Text k=100 \\ \hline
 Concat &53 &52 &52 \\ \hline
 MCB    &58 &57 &57 \\ \hline  
 \end{tabular}
\end{table}

\subsection{VQA Task}
\begin{table}[h]
\small
\caption{Classification accuracy using Fused text and Image Features and question features on VQA task.}\label{vqatextnimage}
 \begin{tabular}{| l |l| l | c |}
 \hline
 Fusion  & Question & Question Image & Question Image Text \\ \hline
 Concat &10.93 &11.9  &12.44 \\ \hline
  
 \end{tabular}
\end{table}

For the VQA task, we observe that using only the question features we can obtain surprising results, but the incorporation of image and text features thereafter, does lead to improved performance.

\section{Conclusion}
Text based semantic embedding, originated from using meta-data, or annotation, when applied to text content from images, are dependant on robust and accurate transcription generation. Thus transcription is a weak link in this pipeline leading to significant performance drops due to vocabulary misses. These vocabulary misses can occur at two levels, misses by the wordspotting engine, and the misses by the word2vec lexicon during vector embedding. To address these issues, we are working towards an end to end embedding scheme, that generates semantic vectors from raw image pixels, without requiring any recognition and transcription.

{\small
\bibliographystyle{ieee}
\bibliography{egbib}
}

\end{document}